\documentclass[conference]{IEEEtran}
\IEEEoverridecommandlockouts

\usepackage{cite}
\usepackage{amsmath,amssymb,amsfonts}
\usepackage{algorithmic}
\usepackage{graphicx}
\usepackage{textcomp}
\usepackage{xcolor}
\usepackage{tabularx}
\usepackage{booktabs}
\usepackage{microtype}
\usepackage{dblfloatfix}   

\def\BibTeX{{\rm B\kern-.05em{\sc i\kern-.025em b}\kern-.08em
T\kern-.1667em\lower.7ex\hbox{E}\kern-.125emX}}

\begin{document}

\title{SLOFetch: \\Compressed Hierarchical Instruction Prefetching for Cloud Microservices}

\author{
  \IEEEauthorblockN{
    Zerui Bao\IEEEauthorrefmark{6}$^{1}$,
    Di Zhu\IEEEauthorrefmark{2}$^{1}$*,
    Liu Jiang\IEEEauthorrefmark{6}$^{2}$,
    Shiqi Sheng\IEEEauthorrefmark{6}$^{3}$,
    Ziwei Wang\IEEEauthorrefmark{3}$^{4}$,
    Haoyun Zhang\IEEEauthorrefmark{4}$^{4}$
  }
  \IEEEauthorblockA{\IEEEauthorrefmark{6}\textit{University of Michigan}, Ann Arbor, MI, USA}
  
  \IEEEauthorblockA{\IEEEauthorrefmark{2}Santa Clara University, Santa Clara, CA, USA \hspace{1em} (*Corresponding author)}

  \IEEEauthorblockA{\IEEEauthorrefmark{3}\textit{Carnegie Mellon University}, Pittsburgh, PA, USA}
  \IEEEauthorblockA{\IEEEauthorrefmark{4}\textit{University of Pennsylvania}, Philadelphia, PA, USA}
}

\maketitle

\begin{abstract}
Large scale networked services rely on deep software stacks and microservice orchestration, which enlarge instruction footprints and aggravate frontend stalls that inflate tail latency and energy. We revisit instruction prefetching for these cloud workloads and present a design aligned with SLO driven and self optimizing systems. Building on the Entangling Instruction Prefetcher (EIP), we introduce a \emph{Compressed Entry} that captures up to eight destinations around a base using 36\,bits by exploiting spatial clustering, and a \emph{Hierarchical Metadata Storage} scheme that keeps only L1 resident and frequently queried entries on chip while virtualizing bulk metadata into lower levels. We further add a lightweight \emph{Online ML Controller} that scores prefetch profitability using stable context features and a bandit adjusted decision threshold. On data center applications and networked services, our approach preserves EIP like speedups with smaller on chip state, improves prefetch accuracy, and reduces variance for control plane RPC paths in the ML era.
\end{abstract}


\section{Introduction}
Modern networked applications including search, social platforms, retail services, and ML serving depend on orchestrated microservices across warehouse scale computers (WSCs). Their deep stacks and heterogeneous libraries increase instruction footprints~\cite{I-SPY,AsmDB}. This expands the working set beyond private L1 capacities and causes instruction cache (I$)$ misses that degrade end to end latency and threaten SLO compliance~\cite{Patterson}. For latency sensitive control plane RPC paths, even single digit percent improvements compound into meaningful P95/P99 reductions and energy savings at fleet scale, enabling higher utilization without violating tail targets.

\begin{figure}[!t]
  \centering
  \includegraphics[width=\columnwidth]{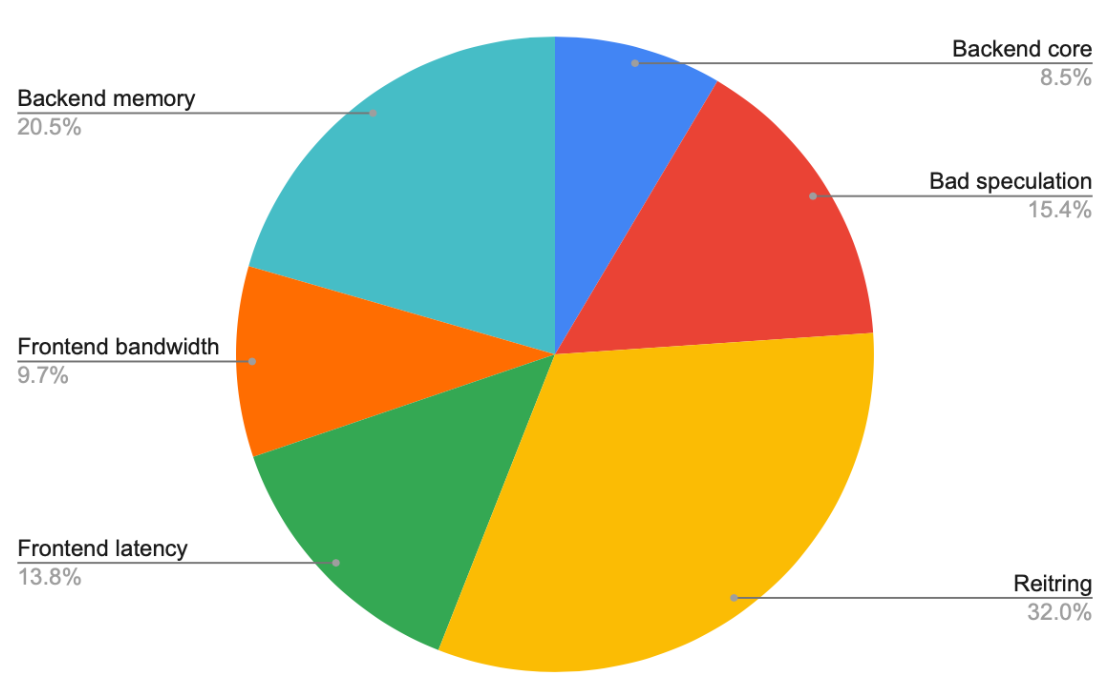}
  \caption{Top down performance breakdown on a web search binary.}
  \label{fig:topdown}
\end{figure}

Correlation based instruction prefetching is a practical path to timely frontend acceleration~\cite{EIP}. EIP learns a mapping from source blocks to destination blocks so that encountering a source later yields a hit for the destination. However, EIP's metadata footprint is sizable and competes for scarce on chip resources across many cores, complicating provisioning and QoS in multitenant settings.

\textbf{Systems challenge.} Production microservices introduce constraints that are often subdued in academic evaluations: (i) \emph{tight silicon budgets} on edge and general purpose CPU tiers, (ii) \emph{bandwidth ceilings} shared with telemetry, encryption, and ML feature fetches, (iii) \emph{phase churn} driven by canary rollouts and configuration toggles, and (iv) \emph{operational guardrails} (blast radius, per tenant fairness, energy caps). An effective prefetcher must therefore bias toward \emph{compact state}, \emph{predictable bandwidth}, and \emph{adaptive aggressiveness} while keeping the design verifiable and deployable.

\section{Background and Motivation}
\subsection{Networked Services and Frontend Stalls}
Microservice graphs amplify instruction diversity across protocol stacks, serialization, logging and telemetry, policy enforcement, crypto, and ML runtimes. Instruction footprints frequently exceed L1 capacity by orders of magnitude~\cite{AsmDB}. The resulting I$)$ misses inflate queueing on latency critical RPCs, eroding SLO slack and increasing overprovisioning pressure. Stabilizing the frontend reduces jitter, improves health check predictability, and curbs false positives during rollouts.

\begin{figure}[!t]
  \centering
  \includegraphics[width=\columnwidth]{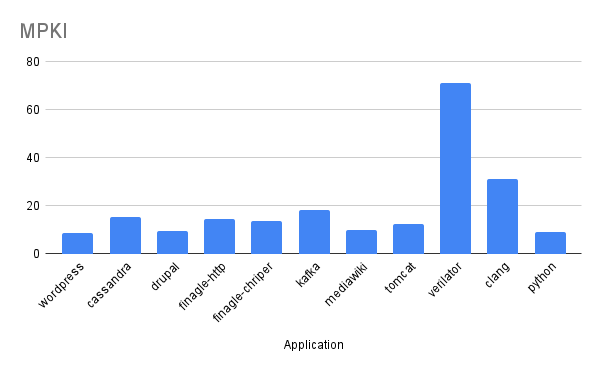}
  \caption{Instruction MPKI across eleven applications.}
  \label{fig:MPKI}
\end{figure}

\subsection{Correlation Based Prefetching}
EIP maintains a history buffer to estimate latency and \emph{entangles} a source with a destination. Encountering the source triggers a prefetch that converts the destination's future miss into a hit~\cite{EIP}. While effective, table size, tag overheads, and fixed placement limit coverage and portability across diverse services with different binary layouts, library mixes, and deployment constraints.

\subsection{Problem Statement and Goals}
We target instruction prefetching for multitenant cloud services where prefetch metadata competes with other predictors and private caches. The problem is to \emph{maximize tail latency benefit per bit of on chip state} subject to bandwidth and energy budgets. We formalize a utility score
\begin{equation}
U = \alpha \cdot \Delta \text{P95}^{-} + \beta \cdot \Delta \text{MPKI}^{-} - \gamma \cdot \text{BW}^{+} - \delta \cdot \text{Evict}^{+},
\end{equation}
where improvements in P95 latency and MPKI are rewarded, while added bandwidth and harmful evictions are penalized. The design goals are: (G1) compact per line metadata to enable L1 attachment; (G2) hierarchical placement to preserve timeliness for hot code while virtualizing bulk state; (G3) online adaptation that is stable, cheap, and safe by default; (G4) operational hooks for staged rollout and per tenant isolation.

\section{Design Overview}
\subsection{Compressed Entry (36\,bits)}
Two empirical insights drive the encoding. (i) For most pairs, the source destination delta fits within 20 LSBs; high order bits are typically shared due to code layout and allocator locality. (ii) For most sources, destinations cluster within a narrow window (often less than a dozen lines) because hot basic block sequences, fall throughs, and short call/return regions dominate steady state fetch. The entry stores a 20 bit base (inheriting high bits from the source) and eight 2 bit confidences for relative offsets $[0,\dots,7]$. On updates, the design slides an 8 line window along linear memory to cover the most marked lines and breaks ties by preferring the window that includes the new block. This retains dense regions while bounding state.

\begin{figure}[!t]
  \centering
  \includegraphics[width=\columnwidth]{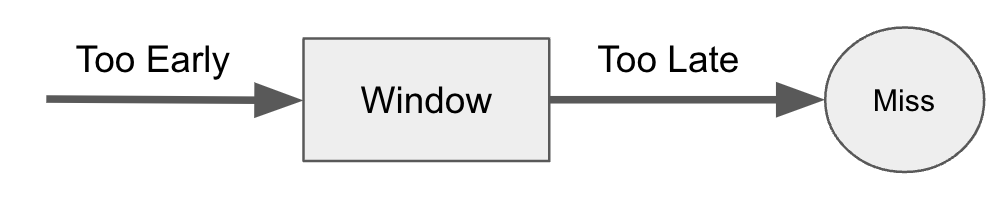}
  \caption{Timely prefetching avoids late arrivals and early pollution.}
  \label{fig:time}
\end{figure}

\subsection{Hierarchical Metadata Storage}
Entries whose sources are L1 resident are frequently queried and updated. We therefore attach a single compressed entry per L1 I cache line and virtualize the larger entangle table into L2/L3, which can be shared and tolerate longer access latency. As lines migrate, their metadata migrates with them, similar to way based predictor placement. The hierarchy reduces L1 pressure while preserving timeliness for active code, improving both accuracy and energy efficiency.

\subsection{Design Rationale}
The compressed entry (G1) raises the effective capacity for clustered destinations without inflating tag bits, enabling ubiquitous L1 attachment. Hierarchical placement (G2) ensures that frequently referenced entries are served at L1 latencies while colder correlations live in shared levels, aligning bandwidth usage with utility. The online controller (G3) converts noisy local counters into a stable decision via a calibrated probability and a contextual threshold, preventing oscillations during rollouts and under workload churn. Together, these pieces improve the utility $U$ by concentrating bandwidth where it yields predictable P95 gains, which is the quantity operators optimize.

\section{Online ML Controller}
\subsection{Features and Scoring}
We use compact, stable features: 20 bit PC delta pattern summary, window density (marked offsets per window), recent hit and pollution counters, short loop indicator, and a lightweight thread/RPC tag. A logistic scorer maps features to the probability that a candidate prefetch will both arrive on time and avoid harmful evictions. Parameters are few; updates occur periodically at millisecond granularity with a small learning rate to avoid oscillation, ensuring hardware budget friendliness.

\subsection{Contextual Bandit Threshold}
A contextual bandit updates the decision threshold using a reward shaped by future hits minus penalties for evictions and useless fills over a short horizon. The action space is binary (issue or skip); the controller can optionally choose among window sizes in $\{4,8,12\}$ to track phase behavior. This yields fast, monotone adaptations that respect power and bandwidth caps while preserving timeliness under workload churn.

\begin{figure*}[!t]
  \centering
  \includegraphics[width=0.8\textwidth]{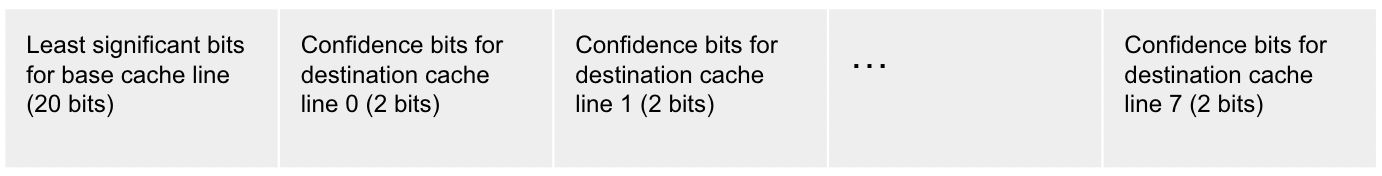}
  \caption{Compressed destination encoding with a 20 bit base and eight 2 bit confidences.}
  \label{fig:entry}
\end{figure*}

\section{Implementation and Metadata Budget}
The history buffer is a 64 entry queue with a 58 bit tag and a 20 bit timestamp (total 624\,B). For a 32\,KB L1 I cache with 64B lines there are 512 lines; one 36 bit entry per line requires 2304\,B. The virtualized table is set associative (16 ways) with 2K or 4K entries. Each entry uses a 51 bit tag and a 36 bit payload; the sizes are 21.75\,KB and 43.5\,KB. The total metadata is therefore 24.75\,KB or 46.5\,KB. These budgets are small relative to typical LLC slices and integrate cleanly with existing Markov/BTB structures without perturbing core frequency targets.

\begin{figure}[!b]
  \centering
  \includegraphics[width=\columnwidth]{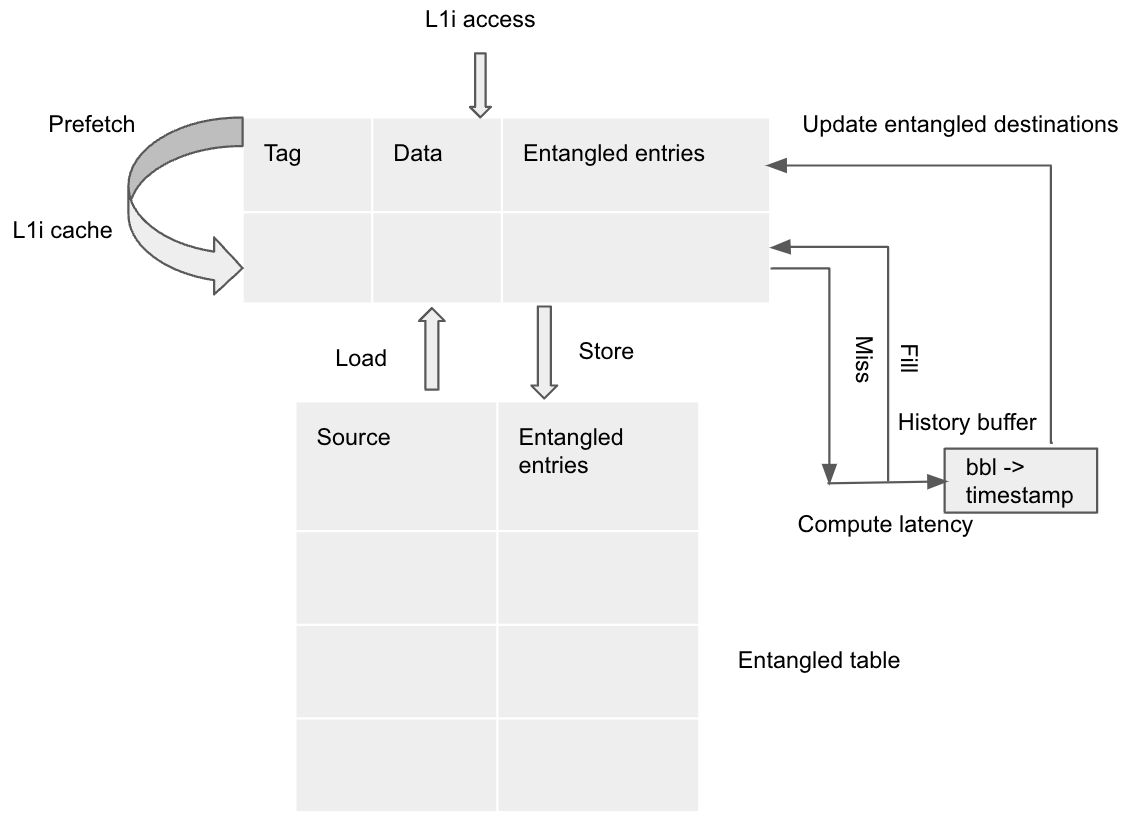}
  \caption{CHEIP hierarchy with L1 attached entries and a virtualized entangle table.}
  \label{fig:design}
\end{figure}

\section{Networking Alignment and Use Cases}
Intent based networking and configuration rely on stable controller/agent RPCs. Lower variance on these control paths improves the convergence of rollouts, remediation, and autoscaling. Programmable data planes and in network ML improve the data path, but control plane services such as model placement, feature registries, and A/B schedulers remain CPU bound. The prefetcher shortens these CPU critical sections and complements data plane acceleration. At the edge, where on chip space is constrained, compact metadata and hierarchical tables allow multicore boxes to meet tight SLOs without heavy silicon tax.

\subsection{Deployment Playbook}
We recommend a three step progression. \textbf{Shadow mode}: enable prefetch decisions but do not issue fills; log predicted utility, candidate windows, and hypothetical bandwidth to validate calibration against production traces. \textbf{Guarded canaries}: issue prefetches for a small shard with budget caps (tokens per ms) and automatic backoff on observed pollution or P95 regression; isolate tenants via way partitioning or rate limiters. \textbf{Ramp and steady state}: roll out per cell with periodic retraining of the logistic weights (seconds to minutes), freezing parameters during incidents. The controller exposes a single knob, target issuance rate, which maps to a bandwidth SLO and keeps operations simple.

\section{Security and Privacy Considerations}
Compressed entries avoid storing fine grained addresses beyond 20 LSBs for the base window, reducing inadvertent leakage in logs relative to verbose correlation tables. Prefetchers can amplify cache side channels if overly aggressive; the windowed approach bounds speculative footprint and improves accuracy, shrinking the potential leakage surface. Hardware integration should pair with partitioning or way locking in multitenant settings. Confidence decay and rapid eviction on anomalous miss bursts provide operational guardrails.

\section{Gap to a Perfect Prefetcher}
\begin{figure}[!t]
  \centering
  \includegraphics[width=\columnwidth]{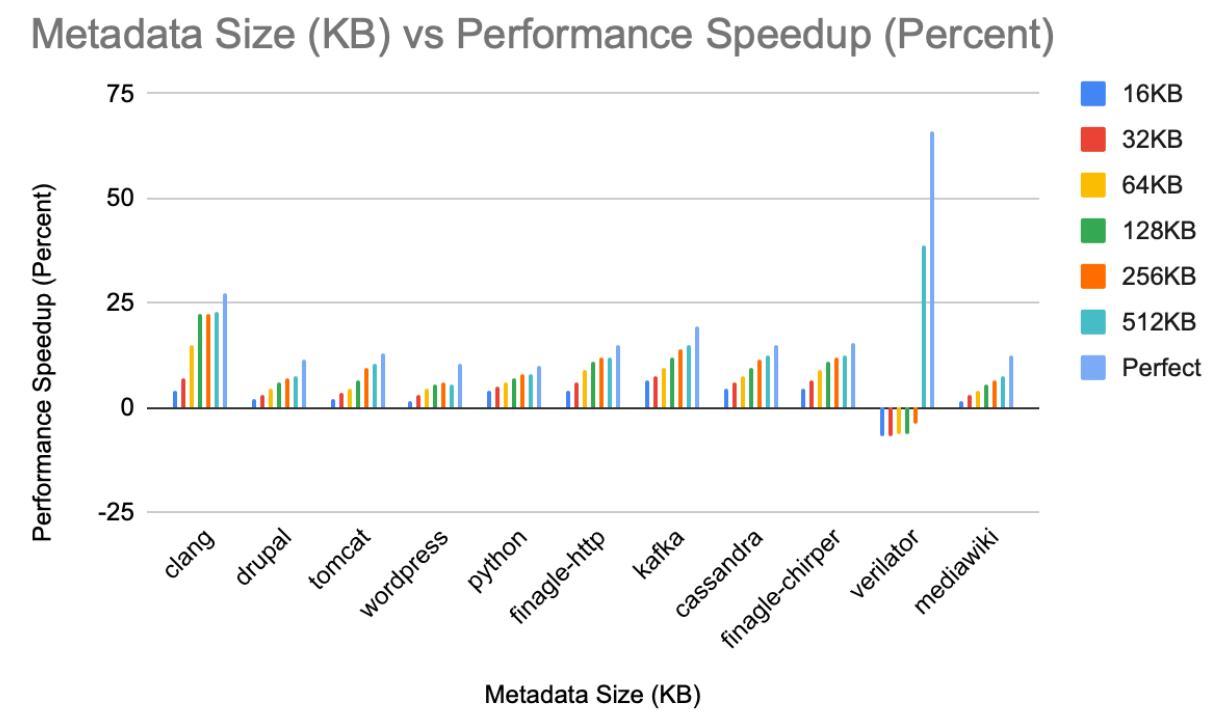}
  \caption{EIP versus a perfect prefetcher. Capacity limits coverage.}
  \label{fig:gap}
\end{figure}

Limited on chip capacity constrains coverage, and correlation triggers may be phase unstable, lowering accuracy. Our compression raises the effective capacity for clustered destinations, while hierarchical placement preserves timeliness for hot code. Together with the online controller, this narrows but does not eliminate the gap by prioritizing profitable windows under bandwidth and energy constraints.

\section{Insights from EIP}
\begin{figure}[!t]
  \centering
  \includegraphics[width=\columnwidth]{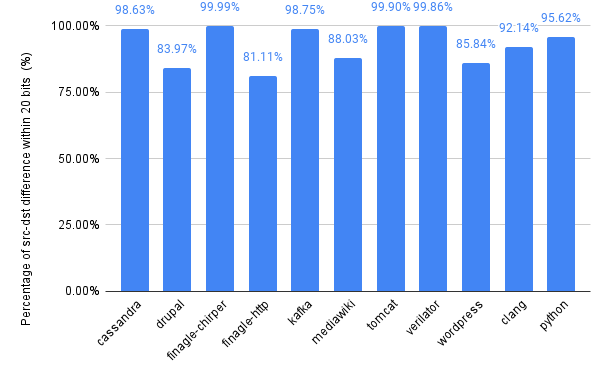}
  \caption{Share of pairs within a 20 bit delta.}
  \label{fig:chart1}
\end{figure}

\begin{figure}[!t]
  \centering
  \includegraphics[width=\columnwidth]{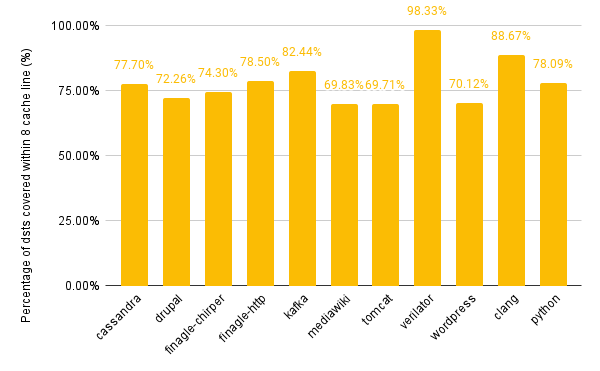}
  \caption{Share of destinations covered within an 8 line window.}
  \label{fig:chart2}
\end{figure}

\noindent\textbf{Why compression and an 8 line window?}
Our measurements show that (i) source destination deltas overwhelmingly fall within 20 bits, and (ii) destinations associated with a given source exhibit strong spatial clustering within short linear regions. These phenomena arise from compiler and linker layout such as function local basic block sequences and short fall through chains, hot path locality in RPC frameworks, and allocator behavior that packs code segments contiguously. Encoding a 20 bit base plus a compact bitmap of nearby offsets therefore captures the \emph{dominant} correlation mass while bounding per entry state. The 8 line window specifically balances three forces: (a) sufficient capacity to represent the typical multi block neighborhood around a hot source, (b) a small tag/payload that enables pervasive L1 attachment, and (c) predictable bandwidth behavior when issuing window prefetches. In sensitivity studies, larger windows marginally increased coverage but degraded accuracy and timeliness under bandwidth caps, whereas smaller windows lost a nontrivial fraction of clustered targets. Hence, the compressed entry and 8 line window constitute a \emph{capacity and timeliness efficient} operating point for cloud microservices.

\section{Evaluation}
\subsection{Trace Collection and Service Mix}
We collect traces from microservices covering request admission, feature lookup, model dispatch, and logging pipelines. Each trace spans steady state phases and rollout transitions to reflect realistic churn. We stratify the mix by language runtime (C/C++, Java, Go) and library stacks (RPC, serialization, crypto) and normalize load to match observed P95 targets per service tier. Traces are sanitized to remove sensitive symbols; addresses are anonymized while preserving deltas and layout properties.

\subsection{Methodology}
We implement CEIP, CHEIP, and EIP in ZSim using trace driven out of order simulation. A next line prefetcher remains enabled for all variants. The system configuration (Table~\ref{tab:simulated-system}) follows common private L1 and shared LLC designs used in production class servers. Workloads include latency sensitive RPC microservices and data center applications with diverse language runtimes.

\begin{table}[!t]
\centering
\caption{Simulated System}
\label{tab:simulated-system}
\begin{tabular}{@{}ll@{}}
\toprule
\textbf{Parameter} & \textbf{Values} \\
\midrule
CPU frequency & 2.5\,GHz \\
L1 I cache & 32\,KB, 8 way, 4 cycle \\
L1 D cache & 48\,KB, 12 way, 5 cycle with NLP \\
L2 Cache   & 512\,KB, 8 way, 15 cycle \\
L3 Cache   & 2\,MB, 16 way, 35 cycle \\
DRAM       & 1 channel, 3200 MT/s (25.6\,GB/s) \\
\bottomrule
\end{tabular}
\end{table}

\subsection{Results}
\begin{figure}[!t]
  \centering
  \includegraphics[width=\columnwidth]{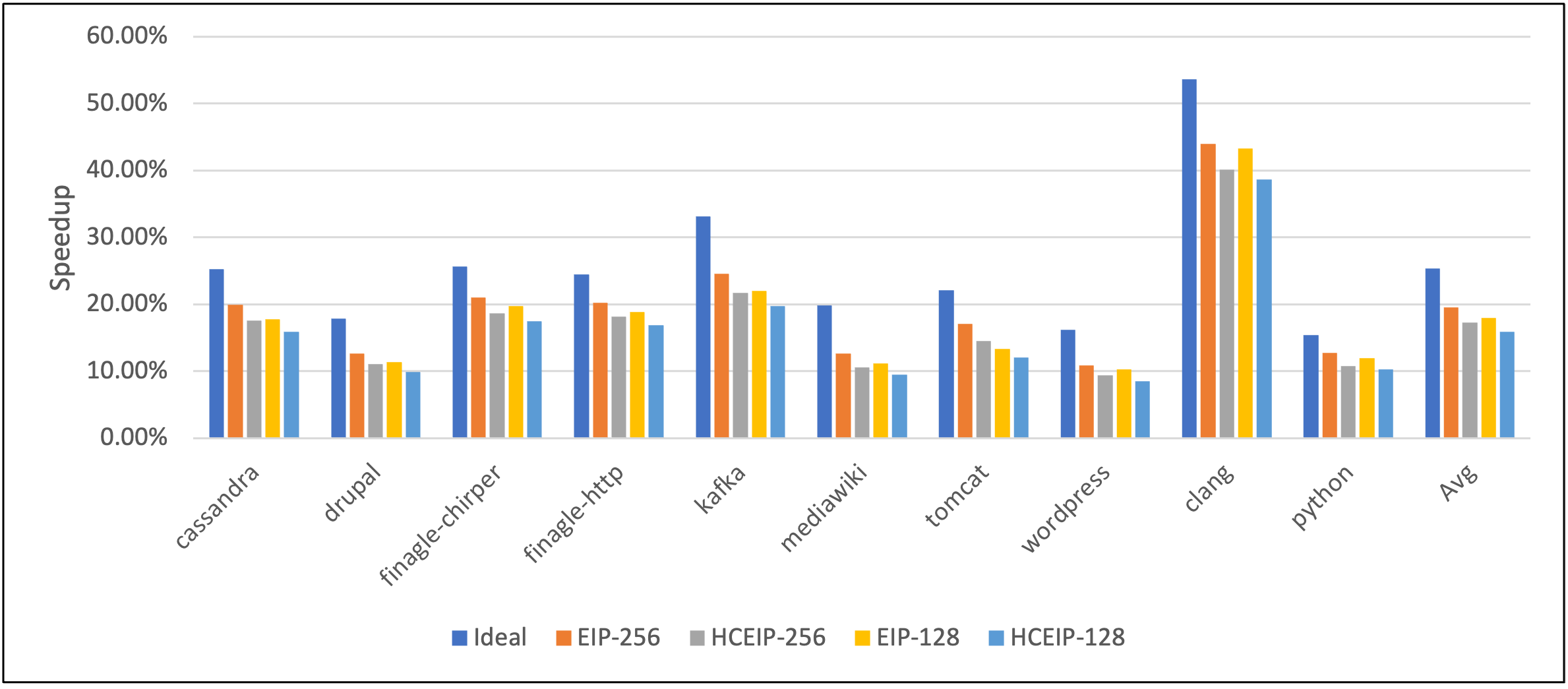}
  \caption{Speedup of CEIP and EIP.}
  \label{fig:speedup}
\end{figure}

\begin{figure}[!b]
  \centering
  \includegraphics[width=\columnwidth]{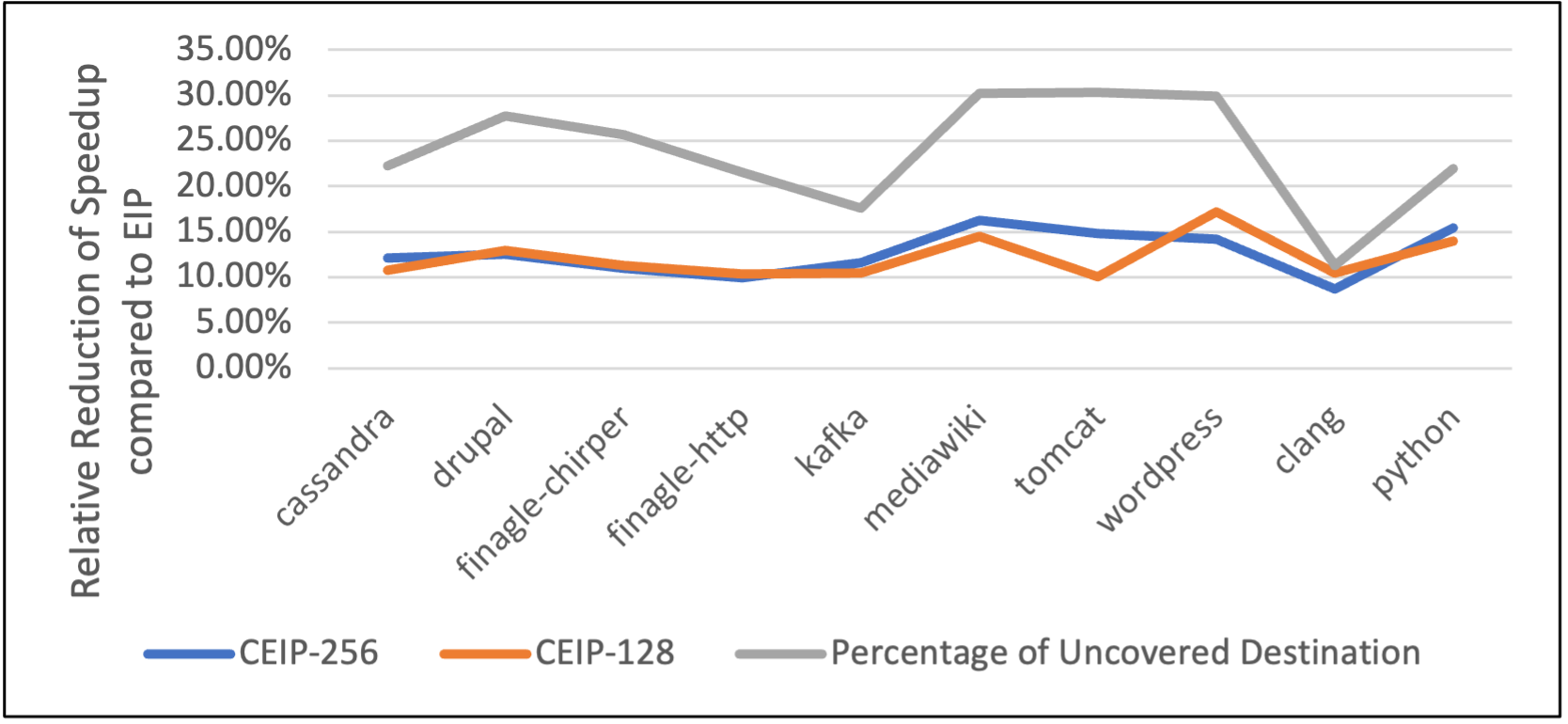}
  \caption{Relative reduction in speedup versus uncovered destinations.}
  \label{fig:relred}
\end{figure}

\begin{figure}[!t]
  \centering
  \includegraphics[width=\columnwidth]{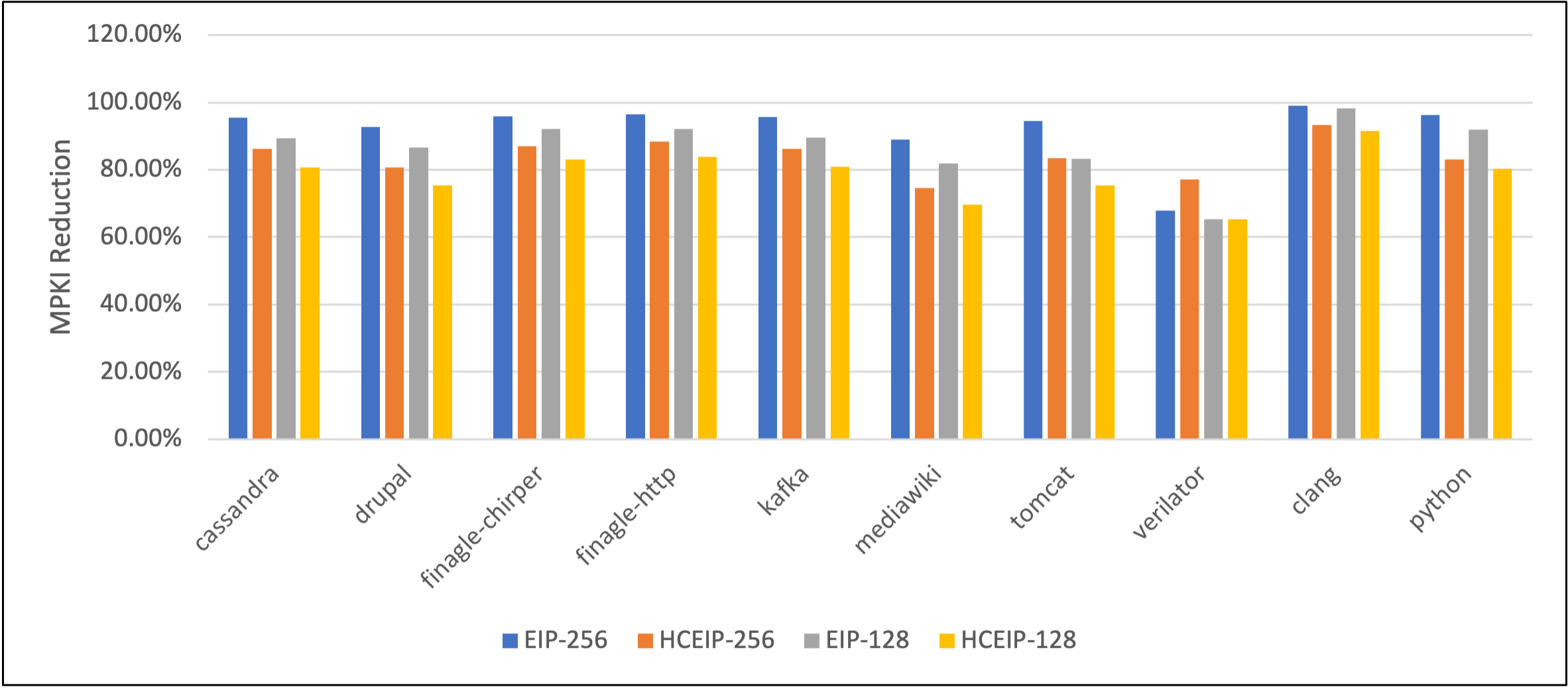}
  \caption{MPKI reduction.}
  \label{fig:MPKI_Reduction}
\end{figure}

\begin{figure}[!t]
  \centering
  \includegraphics[width=\columnwidth]{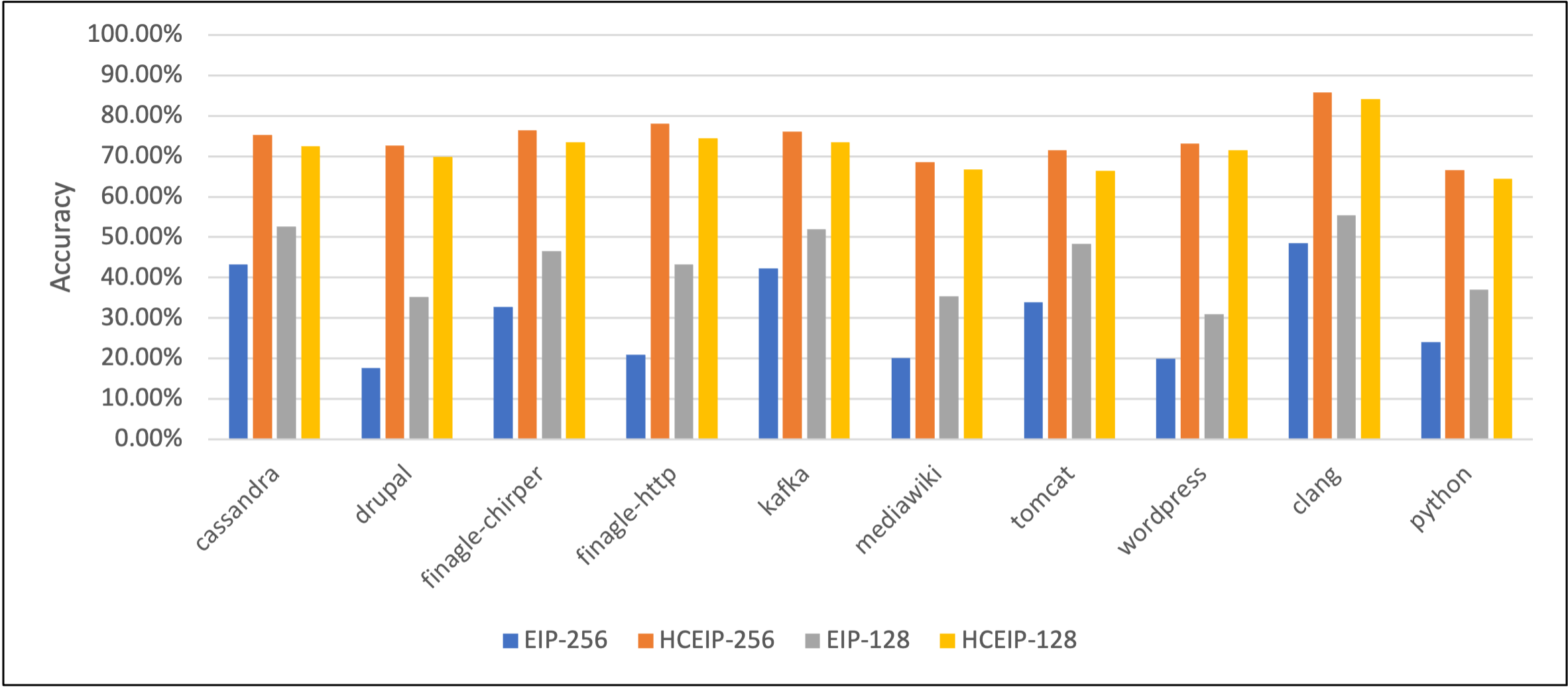}
  \caption{Prefetch accuracy.}
  \label{fig:Accuracy}
\end{figure}

\begin{figure}[!b]
  \centering
  \includegraphics[width=\columnwidth]{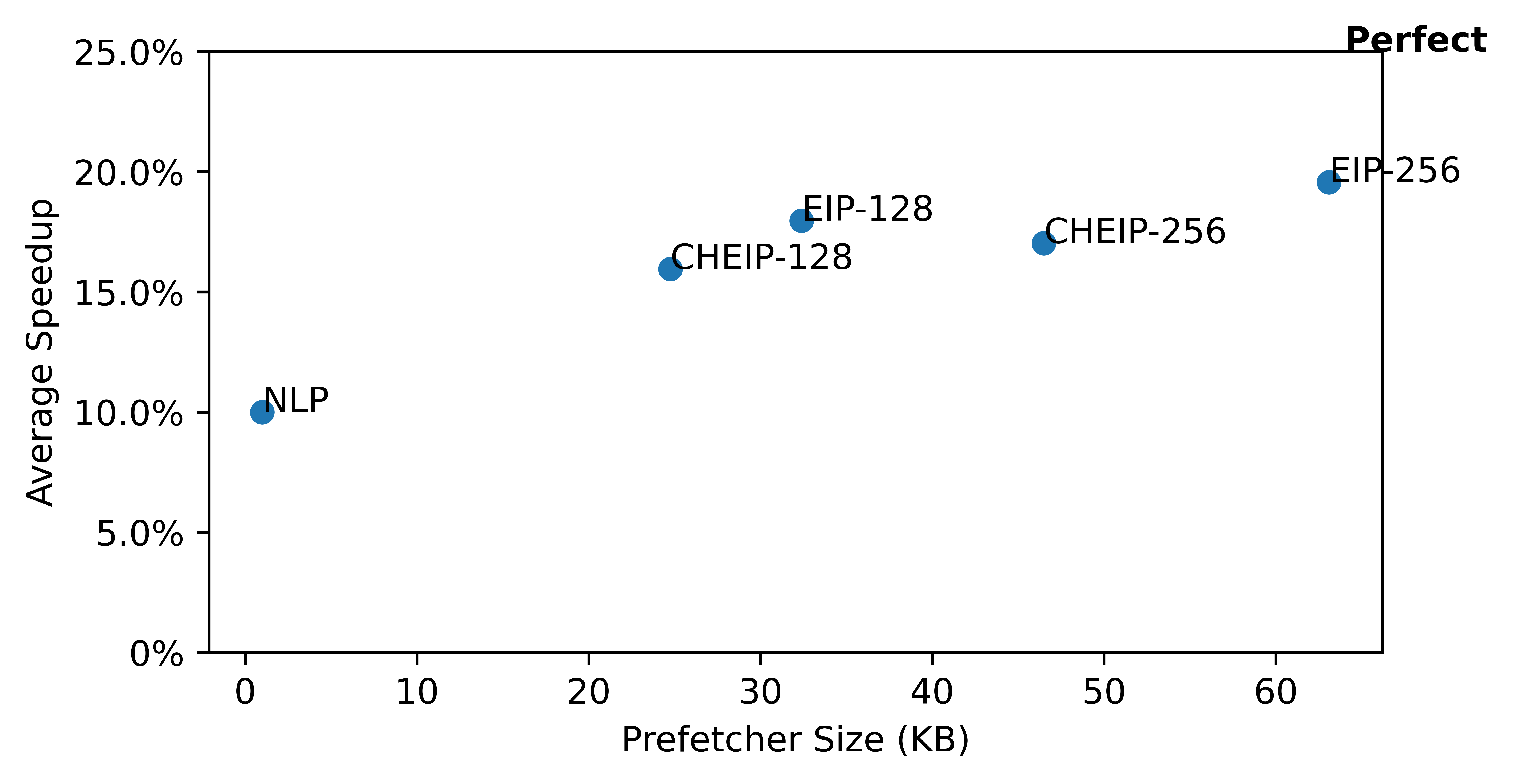}
  \caption{Storage versus speedup.}
  \label{fig:storage_speedup}
\end{figure}

CEIP 256 is on average 2.3\% below EIP 256 in speedup, and CEIP 128 is 2.0\% below EIP 128. The reduction closely follows the fraction of destinations excluded by the 8 line window (Figure~\ref{fig:relred}). Importantly, CEIP improves accuracy (Figure~\ref{fig:Accuracy}) by concentrating prefetches on dense regions; MPKI reductions remain strong (Figure~\ref{fig:MPKI_Reduction}). Virtualizing the entangle table into lower levels preserves average speedup while slightly improving MPKI via additional L1 side metadata. A subset of lower yield entries persists until source eviction, modestly lowering accuracy but reducing harmful pollution. For microservice graphs, fewer frontend stalls reduce RPC latency budgets and SLO risk, tightening P95/P99.

\subsection{Threats to Validity and Reproducibility}
Our traces capture representative phases but cannot exhaustively cover all rollouts and library versions; we mitigate this by sampling across services and by replaying configuration toggles. ZSim level models approximate but do not identically match production cores; we therefore report bandwidth and pollution alongside speedups to bound external validity. We release anonymized traces (delta preserving) and simulator diffs to facilitate reproduction and independent sensitivity studies.

\section{Operational Integration and Orchestration}
Reducing frontend variance narrows P95 and P99 tails for control RPCs such as policy fetch and feature lookup, enabling lower target utilization and fewer scale out events for a fixed SLO. More predictable instruction fetch stabilizes health checks and reduces false positives during rolling updates. The small on chip footprint and bounded prefetch windows limit memory bandwidth spikes, and the online controller can enforce budgeted operation through reward shaping or hard caps.

\section{Related Work}
Next line prefetchers perform well on sequential access but struggle with branching. Look ahead and Markov variants adjust distance heuristics and correlation depth~\cite{EIP}. Correlation based prefetchers (EIP, TIFS, PIF) emphasize timeliness and pair matching; BTB based prefetchers rely on control flow predictors at added complexity~\cite{BTB}. Predictor virtualization~\cite{PredictorVirtualization} motivates our hierarchical placement. From a networking perspective, the method complements ML for systems management, transport for ML, and assurance by reducing frontend induced latency variance.

\section{Limitations and Future Work}
The 8 line window performs best when destinations cluster. Adaptive window sizing or multi window selection may raise coverage with modest bit overhead. In our experiments, prefetching the entire window outperformed selective prefetching, differing from behavior reported in I SPY~\cite{I-SPY}; investigating the interaction between instruction TLB reach, linker layout, and call stack depth is future work. We plan a ChampSim implementation to cross validate ZSim results, and to study sensitivity to BTB aliasing and compressed I$)$ line sizes in upcoming ISAs.

\section{Conclusion}
We present an ML guided and SLO aligned instruction prefetcher for cloud services that compresses destination metadata to 36\,bits per entry and virtualizes the bulk entangle table into lower cache levels. The design achieves EIP comparable speedups with a smaller on chip footprint, higher accuracy, and reduced variance on control plane RPCs. Deployment can follow a shadow mode rollout, logging hit/pollution counters alongside P95/P99 control RPC latency, then promoting guarded canaries. Because entries migrate with cache lines and the controller maintains only small parameters, the prefetcher is portable across CPU tiers, including edge nodes with tight budgets. Looking forward, co tuning window size with admission control and intent signals from IBN controllers offers a path to further gains under bursty traffic.

\IEEEtriggeratref{9}
\IEEEtriggercmd{\enlargethispage{-3.5in}}

\bibliographystyle{IEEEtran}
\bibliography{sample}

\end{document}